\documentclass{article}
\usepackage{times}
\usepackage{graphicx} 
\usepackage{subfigure}
\usepackage{natbib}
\usepackage{algorithm}
\usepackage{algorithmic}
\usepackage{hyperref}

\usepackage[accepted]{icml2017}
\usepackage{url}
\usepackage{latexsym}
\usepackage{tabularx}
\usepackage{multirow}
\usepackage{amsmath}
\usepackage[normalem]{ulem}
\usepackage{epsfig}
\usepackage{amsmath}
\usepackage{amsfonts}
\usepackage{amsthm}
\usepackage{amssymb}

\usepackage[group-separator={,},group-minimum-digits={3}]{siunitx}

\newcommand{\specialcell}[3]{\begin{tabular}[#1]{@{}#2@{}}#3\end{tabular}}

\icmltitlerunning{Sparsified Back Propagation for Accelerated Deep Learning with Reduced Overfitting}

\begin{document}

\twocolumn[
\icmltitle{meProp: Sparsified Back Propagation for Accelerated Deep Learning\\ with Reduced Overfitting}

\icmlsetsymbol{equal}{*}

\begin{icmlauthorlist}
\icmlauthor{Xu Sun}{eecs,moe}
\icmlauthor{Xuancheng Ren}{eecs,moe}
\icmlauthor{Shuming Ma}{eecs,moe}
\icmlauthor{Houfeng Wang}{eecs,moe}
\end{icmlauthorlist}

\icmlaffiliation{eecs}{School of Electronics Engineering and Computer Science, Peking University, China}
\icmlaffiliation{moe}{MOE Key Laboratory of Computational Linguistics, Peking University, China}

\icmlcorrespondingauthor{Xu Sun}{xusun@pku.edu.cn}

\icmlkeywords{acceleration, sparse learning, neural networks, overfitting}

\vskip 0.3in
]

\printAffiliationsAndNotice{} 

\begin{abstract}

We propose a simple yet effective technique for neural network learning. The forward propagation is computed as usual. In back propagation, only a small subset of the full gradient is computed to update the model parameters. The gradient vectors are sparsified in such a way that only the top-$k$ elements (in terms of magnitude) are kept. As a result, only $k$ rows or columns (depending on the layout) of the weight matrix are modified, leading to a linear reduction ($k$ divided by the vector dimension) in the computational cost. Surprisingly, experimental results demonstrate that we can update only 1--4\% of the weights at each back propagation pass. This does not result in a larger number of training iterations. More interestingly, the accuracy of the resulting models is actually improved rather than degraded, and a detailed analysis is given. The code is available at \url{https://github.com/lancopku/meProp}.

\end{abstract}

\section{Introduction}

Neural network learning is typically slow, where back propagation usually dominates the computational cost during the learning process.
Back propagation entails a high computational cost because it needs to compute full gradients and update all model parameters in each learning step. It is not uncommon for a neural network to have a massive number of model parameters.

In this study, we propose a \emph{minimal effort} back propagation method, which we call \emph{meProp}, for neural network learning. The idea is that we compute only a very small but critical portion of the gradient information, and update only the corresponding minimal portion of the parameters in each learning step. This leads to sparsified gradients, such that only highly relevant parameters are updated and other parameters stay untouched. The sparsified back propagation leads to a linear reduction in the computational cost.

To realize our approach, we need to answer two questions.
The first question is how to find the highly relevant subset of the parameters from the current sample in stochastic learning. We propose a top-$k$ search method to find the most important parameters. Interestingly, experimental results demonstrate that we can update only 1--4\% of the weights at each back propagation pass. This does not result in a larger number of training iterations.
The proposed method is general-purpose and it is independent of specific models and specific optimizers (e.g., Adam and AdaGrad).

The second question is whether or not this minimal effort back propagation strategy will hurt the accuracy of the trained models. We show that our strategy does not degrade the accuracy of the trained model, even when a very small portion of the parameters is updated. More interestingly, our experimental results reveal that our strategy actually improves the model accuracy in most cases.
Based on our experiments, we find that it is probably because the minimal effort update does not modify weakly relevant parameters in each update, which makes overfitting less likely, similar to the \emph{dropout} effect.

\begin{figure*}[t]
\begin{center}
\begin{tabular}{c}
    \includegraphics[width=0.8\hsize]{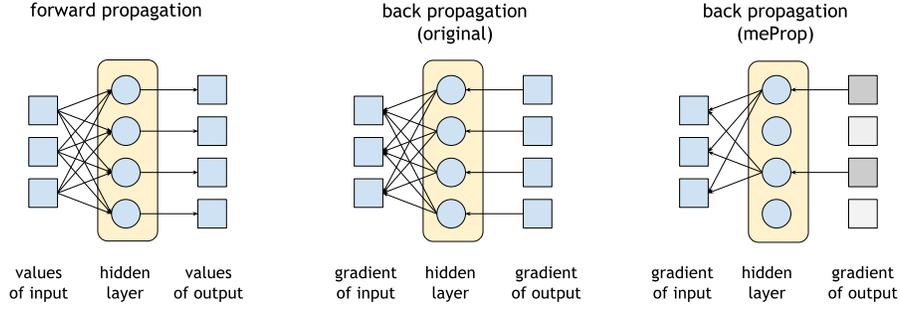}
\end{tabular}
\caption{An illustration of meProp.
}\label{fig1}
\end{center}
\vspace{-0.15in}
\end{figure*}

The contributions of this work are as follows:

\begin{itemize}
\item
We propose a sparsified back propagation technique for neural network learning, in which only a small subset of the full gradient is computed to update the model parameters. Experimental results demonstrate that we can update only 1--4\% of the weights at each back propagation pass. This does not result in a larger number of training iterations.

\item Surprisingly, our experimental results reveal that the accuracy of the resulting models is actually improved, rather than degraded. We demonstrate this effect by conducting experiments on different deep learning models (LSTM and MLP), various optimization methods (Adam and AdaGrad), and diverse tasks (natural language processing and image recognition).

\end{itemize}

\begin{figure*}[t]
\begin{center}
\begin{tabular}{c}
    \includegraphics[width=0.75\hsize]{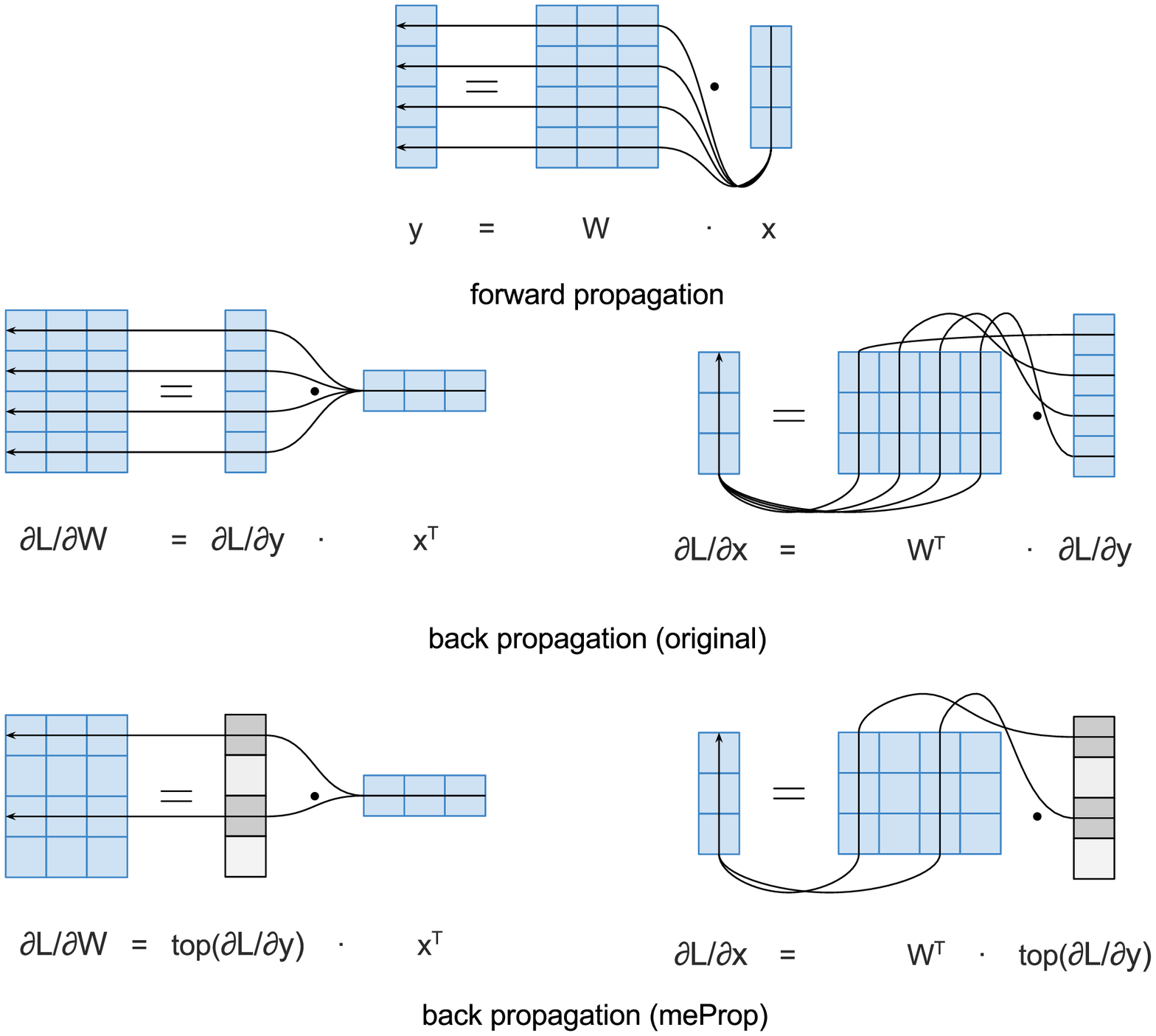}
\end{tabular}
\caption{An illustration of the computational flow of meProp.
}\label{fig2}
\end{center}
\vspace{-0.15in}
\end{figure*}

\begin{figure*}[t]
\begin{center}
\begin{tabular}{c}
    \includegraphics[width=0.75\hsize]{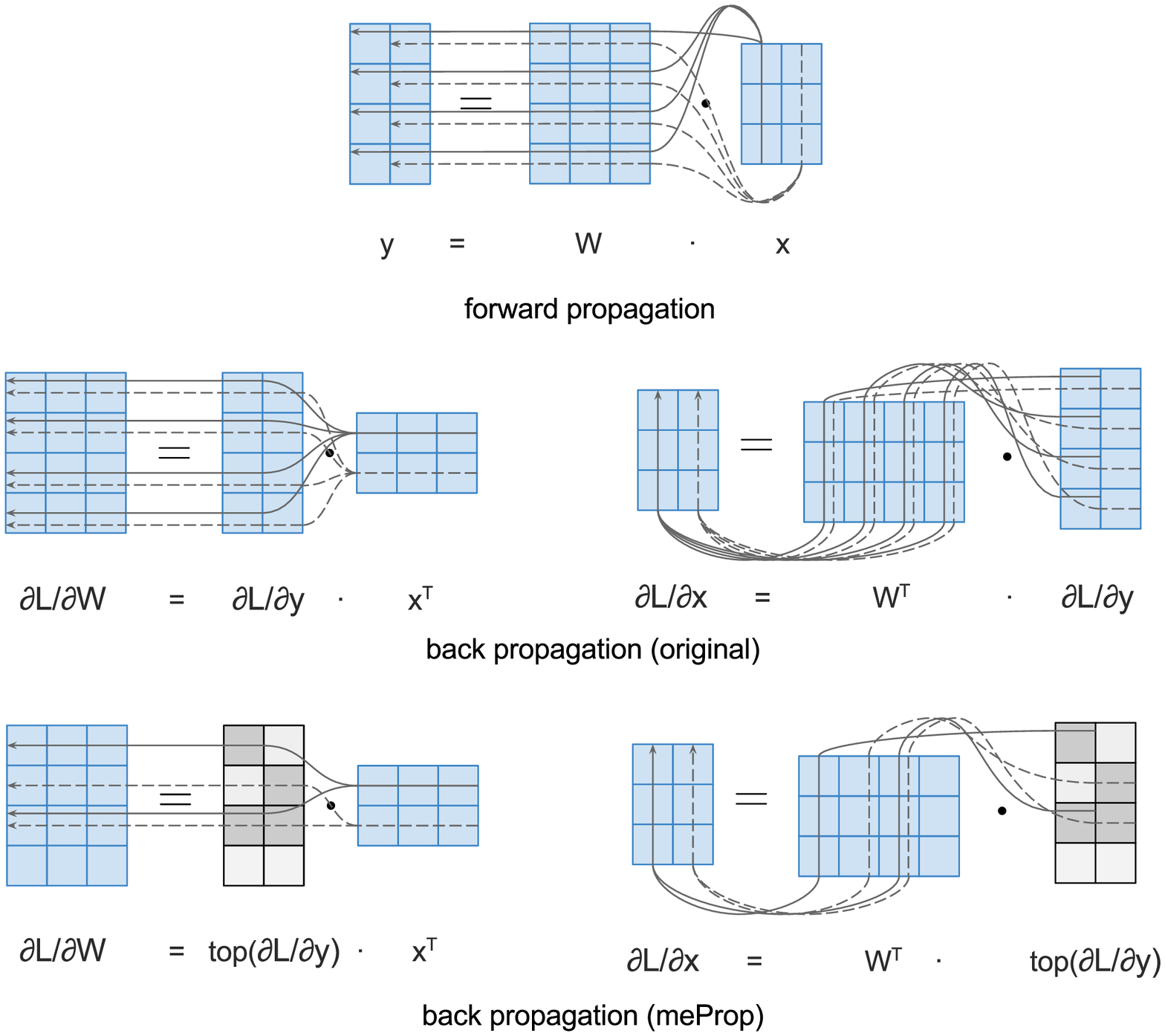}
\end{tabular}
\caption{An illustration of the computational flow of meProp on a mini-batch learning setting.
}\label{fig3}
\end{center}
\vspace{-0.15in}
\end{figure*}

\section{Proposed Method}

We propose a simple yet effective technique for neural network learning. The forward propagation is computed as usual. During back propagation, only a small subset of the full gradient is computed to update the model parameters. The gradient vectors are ``quantized'' so that only the top-$k$ components in terms of magnitude are kept.
We first present the proposed method and then describe the implementation details.

\subsection{meProp}

Forward propagation of neural network models, including feedforward neural networks, RNN, LSTM, consists of linear transformations and non-linear
transformations. For simplicity, we take a computation unit with one linear
transformation and one non-linear transformation as an example:
\begin{equation}\label{linear}
   y= W  x
\end{equation}
\begin{equation}\label{nonlinear}
   z=\sigma(y)
\end{equation}
where $W \in R^{n \times m},  x \in R^{m},  y \in R^{n},  z \in R^{n}$, $m$ is the dimension of the input vector, $n$ is the dimension of the output vector, and $\sigma$ is a non-linear function (e.g., \emph{relu}, \emph{tanh}, and \emph{sigmoid}). During back propagation, we need to compute the gradient of the parameter matrix $W$ and the input vector $x$:
\begin{equation}\label{der1}
   \frac{\partial  z}{\partial  W_{ij}}=\sigma^{'}_{i}{ x}^{T}_{j} \ \ \ (1\leq i \leq n, 1\leq j \leq m)
\end{equation}
\begin{equation}\label{der2}
   \frac{\partial  z}{\partial  x_{i}}=\sum_{j}{{ W}^{T}_{ij}\sigma^{'}_{j}} \ \ \ (1\leq j\leq n, 1\leq i \leq m)
\end{equation}
where $\sigma^{'}_i \in R^{n}$ means $\frac{\partial  z_i}{\partial  y_i}$. We can see that the computational cost of back propagation is directly proportional to the dimension of output vector $n$.

The proposed meProp uses approximate gradients by keeping only top-$k$ elements based on the \emph{magnitude values}. That is, only the top-$k$ elements with the largest absolute values are kept. For example, suppose a vector $v=\langle 1,2,3,-4 \rangle$, then $top_2(v)= \langle 0,0,3,-4 \rangle$.
We denote the indices of vector $\sigma^{'}(y)$'s top-$k$ values as $\{t_1, t_2, ..., t_k\}(1 \leq k \leq n)$, and the approximate gradient of the parameter matrix $W$ and input vector $x$ is:
\begin{equation}\label{der3}
   \frac{\partial z}{\partial W_{ij}} \leftarrow \sigma^{'}_{i}x^{T}_{j} \ \  \text{ if } \ \ i \in \{t_1, t_2, ..., t_k\} \ \ \text{ else } \ \ 0
\end{equation}
\begin{equation}\label{der4}
   \frac{\partial z}{\partial x_{i}} \leftarrow \sum_{j}{W^{T}_{ij}\sigma^{'}_{j}} \ \  \text{ if } \ \ j \in \{t_1, t_2, ..., t_k\} \ \ \text{ else } \ \ 0
\end{equation}
As a result, only $k$ rows or columns (depending on the layout) of the weight matrix are modified, leading to a linear reduction ($k$ divided by the vector dimension) in the computational cost.

Figure~\ref{fig1} is an illustration of meProp for a single computation unit of neural models. The original back propagation uses the full gradient of the output vectors to compute the gradient of the parameters. The proposed method selects the top-$k$ values of the gradient of the output vector, and backpropagates the loss through the corresponding subset of the total model parameters.

As for a complete neural network framework with a loss $L$, the original back propagation computes the gradient of the parameter matrix $W$ as:
\begin{equation}
    \frac{\partial L}{\partial W} = \frac{\partial L}{\partial y} \cdot \frac{\partial y}{\partial W}
\end{equation}
while the gradient of the input vector $x$ is:
\begin{equation}
    \frac{\partial L}{\partial x} = \frac{\partial y}{\partial x} \cdot \frac{\partial L}{\partial y}
\end{equation}
The proposed meProp selects top-$k$ elements of the gradient $\frac{\partial L}{\partial y}$ to approximate the original gradient, and passes them through the gradient computation graph according to the chain rule. Hence, the gradient of $W$ goes to:
\begin{equation}
    \frac{\partial L}{\partial W} \leftarrow  top_k(\frac{\partial L}{\partial y}) \cdot \frac{\partial y}{\partial W}
\end{equation}
while the gradient of the vector $x$ is:
\begin{equation}
    \frac{\partial L}{\partial x} \leftarrow  \frac{\partial y}{\partial x} \cdot top_k(\frac{\partial L}{\partial y})
\end{equation}

Figure~\ref{fig2} shows an illustration of the computational flow of meProp. The forward propagation is the same as traditional forward propagation, which computes the output vector via a matrix multiplication operation between two input tensors. The original back propagation computes the full gradient for the input vector and the weight matrix. For meProp, back propagation computes an approximate gradient by keeping top-$k$ values of the backward flowed gradient and masking the remaining values to 0.

Figure~\ref{fig3} further shows the computational flow of meProp for the mini-batch case.

\subsection{Implementation}\label{sec:imp}

We have coded two neural network models, including an LSTM model for part-of-speech (POS) tagging, and a feedforward NN model (MLP) for transition-based dependency parsing and MNIST image recognition. We use the optimizers with automatically adaptive learning rates, including Adam~\cite{Kingma2014} and AdaGrad~\cite{Duchi2011}. In our implementation, we make no modification to the optimizers, although there are many zero elements in the gradients.

Most of the experiments on CPU are conducted on the framework coded in C\# on our own. This framework builds a dynamic computation graph of the model for each sample, making it suitable for data in variable lengths. A typical training procedure contains four parts: building the computation graph, forward propagation, back propagation, and parameter update. We also have an implementation based on the \emph{PyTorch} framework for GPU based experiments.

\subsubsection{Where to apply meProp}

The proposed method aims to reduce the complexity of the back propagation by reducing the elements in the computationally intensive operations. In our preliminary observations, matrix-matrix or matrix-vector multiplication consumed more than 90\% of the time of back propagation. In our implementation, we apply meProp only to the back propagation from the output of the multiplication to its inputs. For other element-wise operations (e.g., activation functions), the original back propagation procedure is kept, because those operations are already fast enough compared with matrix-matrix or matrix-vector multiplication operations.

If there are multiple hidden layers, the top-$k$ sparsification needs to be applied to every hidden layer, because the sparsified gradient will again be dense from one layer to another. That is, in meProp the gradients are sparsified with a top-$k$ operation at the output of every hidden layer.

While we apply meProp to all hidden layers using the same $k$ of top-$k$, usually the $k$ for the output layer could be different from the $k$ for the hidden layers, because the output layer typically has a very different dimension compared with the hidden layers. For example, there are 10 tags in the MNIST task, so the dimension of the output layer is 10, and we use an MLP with the hidden dimension of 500. Thus, the best $k$ for the output layer could be different from that of the hidden layers.

\subsubsection{Choice of top-$k$ algorithms}

Instead of sorting the entire vector, we use the well-known min-heap based top-$k$ selection method, which is slightly changed to focus on memory reuse. The algorithm has a time complexity of $O(n \log k)$ and a space complexity of $O(k)$.


\begin{table*}[t]
  \centering
  \footnotesize
  \caption{Results based on LSTM/MLP models and AdaGrad/Adam optimizers. Time means averaged time per iteration. Iter means the number of iterations to reach the optimal score on development data. The model of this iteration is then used to obtain the test score.} \label{tab1}
  \begin{tabular}{|l|r r l l|}
    \hline
    \textbf{POS-Tag (AdaGrad)} & Iter & Backprop time (s) & Dev Acc (\%) & Test Acc (\%) \\
    \hline
     LSTM (h=500) & 4 & \num{17534.4}\hspace{28.5pt} &  96.89 & 96.93    \\

     meProp (k=5) & 4 & \textbf{253.2 (69.2x)} & 97.18 (+0.29) & \textbf{97.25 (+0.32)} \\
    \hline

    \textbf{Parsing (AdaGrad)} & Iter & Backprop time (s) & Dev UAS (\%) & Test UAS (\%) \\
    \hline
     MLP (h=500) & 11 & \num{8899.7}\hspace{28.5pt} & 89.07 & 88.92   \\

     meProp (k=20) & 8 & \textbf{492.3 (18.1x)} & 89.17 (+0.10) & \textbf{88.95 (+0.03)} \\
    \hline

    \textbf{MNIST (AdaGrad)} & Iter & Backprop time (s) & Dev Acc (\%) & Test Acc (\%) \\
    \hline
     MLP (h=500) & 8 & \num{171.0}\hspace{28.5pt} & 98.20 & 97.52   \\

     meProp (k=10) & 16 & \textbf{4.1 (41.7x)} & 98.20 (+0.00) & \textbf{98.00 (+0.48)} \\
    \hline
    \hline

    \textbf{POS-Tag (Adam)} & Iter & Backprop time (s) & Dev Acc (\%) & Test Acc (\%)\\
    \hline
     LSTM (h=500) & 2  & \num{16167.3}\hspace{28.5pt}   & 97.18 & 97.07 \\

     meProp (k=5) & 5 & \textbf{247.2 (65.4x)} & 97.14 (-0.04) & \textbf{97.12 (+0.05)} \\
    \hline

    \textbf{Parsing (Adam)} & Iter & Backprop time (s) & Dev UAS (\%) & Test UAS (\%)\\
    \hline
     MLP (h=500) & 14 &  \num{9077.7}\hspace{28.5pt}  & 90.12 & 89.84 \\

     meProp (k=20) & 6 & \textbf{488.7 (18.6x)} & 90.02 (-0.10) &  \textbf{90.01 (+0.17)} \\
    \hline

    \textbf{MNIST (Adam)} & Iter & Backprop time (s) & Dev Acc (\%) & Test Acc (\%)\\
    \hline
     MLP (h=500) &  17  & \num{169.5}\hspace{28.5pt}   & 98.32 & 97.82 \\

     meProp (k=20) & 15 & \textbf{7.9 (21.4x)} & 98.22 (-0.10) & \textbf{98.01 (+0.19)} \\
    \hline
  \end{tabular}
  \vspace{-0.05in}
\end{table*}

\begin{table}[t]
  \centering
  \footnotesize
  \caption{Overall forward propagation time vs. overall back propagation time. Time means averaged time per iteration. FP means forward propagation. BP means back propagation. Ov. time means overall training time (FP + BP).} \label{tab2}
  \begin{tabular}{|l@{}|r@{}|r@{}|r@{}|}
    \hline

    \textbf{POS-Tag (Adam)} & \specialcell{c}{c}{Ov. FP time} & \specialcell{c}{c}{Ov. BP time}\hspace{8.5pt} & \specialcell{c}{c}{Ov. time} \\
    \hline
     LSTM (h=500) & \num{7334}s &   \num{16522}s\hspace{28.5pt}  & \num{23856}s \\

     meProp (k=5) & \num{7362}s & \textbf{540s (30.5x)} & \textbf{7,903s} \\
    \hline

    \textbf{Parsing (Adam)} & \specialcell{c}{c}{Ov. FP time} & \specialcell{c}{c}{Ov. BP time}\hspace{8.5pt} & \specialcell{c}{c}{Ov. time} \\
    \hline
     MLP (h=500) & \num{3906}s  & \num{9114}s\hspace{28.5pt}   & \num{13020}s \\

     meProp (k=20) & \num{4002}s  & \textbf{513s (17.7x)}  & \textbf{4,516s} \\
    \hline

    \textbf{MNIST (Adam)} & \specialcell{c}{c}{Ov. FP time} & \specialcell{c}{c}{Ov. BP time}\hspace{8.5pt} & \specialcell{c}{c}{Ov. time} \\

    \hline
     MLP (h=500) & \num{69}s  & \num{171}s\hspace{28.5pt}  & \num{240}s \\

     meProp (k=20) & \num{68}s & \textbf{9s (18.4x)} &  \textbf{77s} \\
    \hline
  \end{tabular}
  \vspace{-0.05in}
\end{table}

\section{Related Work}

Riedmiller and Braun~\yrcite{riedmiller1993} proposed a direct adaptive method for fast learning, which performs a local adaptation of the weight update according to the behavior of the error function. Tollenaere~\yrcite{tollenaere1990} also proposed an adaptive acceleration strategy for back propagation. Dropout~\cite{Srivastava2014} is proposed to improve training speed and reduce the risk of overfitting.
Sparse coding is a class of unsupervised methods for learning sets of over-complete bases to represent data efficiently~\cite{olshausen1996}. Ranzato et al.~\yrcite{Ranzato2006} proposed a sparse autoencoder model for learning sparse over-complete features.
The proposed method is quite different compared with those prior studies on back propagation, dropout, and sparse coding.

The \emph{sampled-output-loss} methods~\cite{JeanEA2015} are limited to the softmax layer (output layer) and are only based on random sampling, while our method does not have those limitations.
The sparsely-gated mixture-of-experts~\cite{ShazeerEA2017} only sparsifies the mixture-of-experts gated layer and it is limited to the specific setting of mixture-of-experts, while our method does not have those limitations.
There are also prior studies focusing on reducing the communication cost in distributed systems~\cite{SeideEA2014,DrydenEA2016}, by quantizing each value of the gradient from 32-bit float to only 1-bit. Those settings are also different from ours.

\section{Experiments}

\begin{figure*}[t]
\centering
\begin{tabular}{@{}c@{}@{}c@{}@{}c@{}@{}c@{}}

\epsfig{file=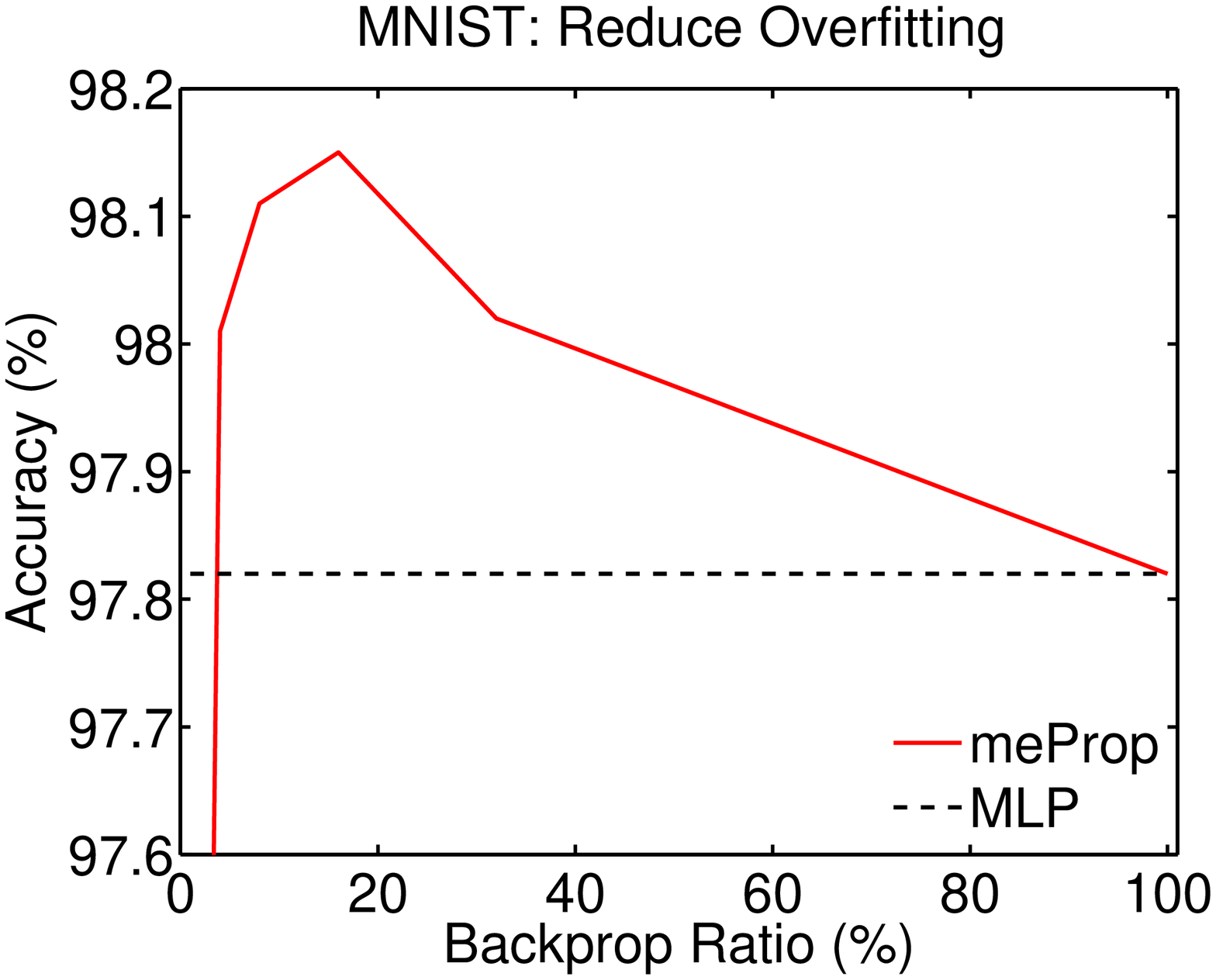,width=0.3\linewidth,clip=}
\epsfig{file=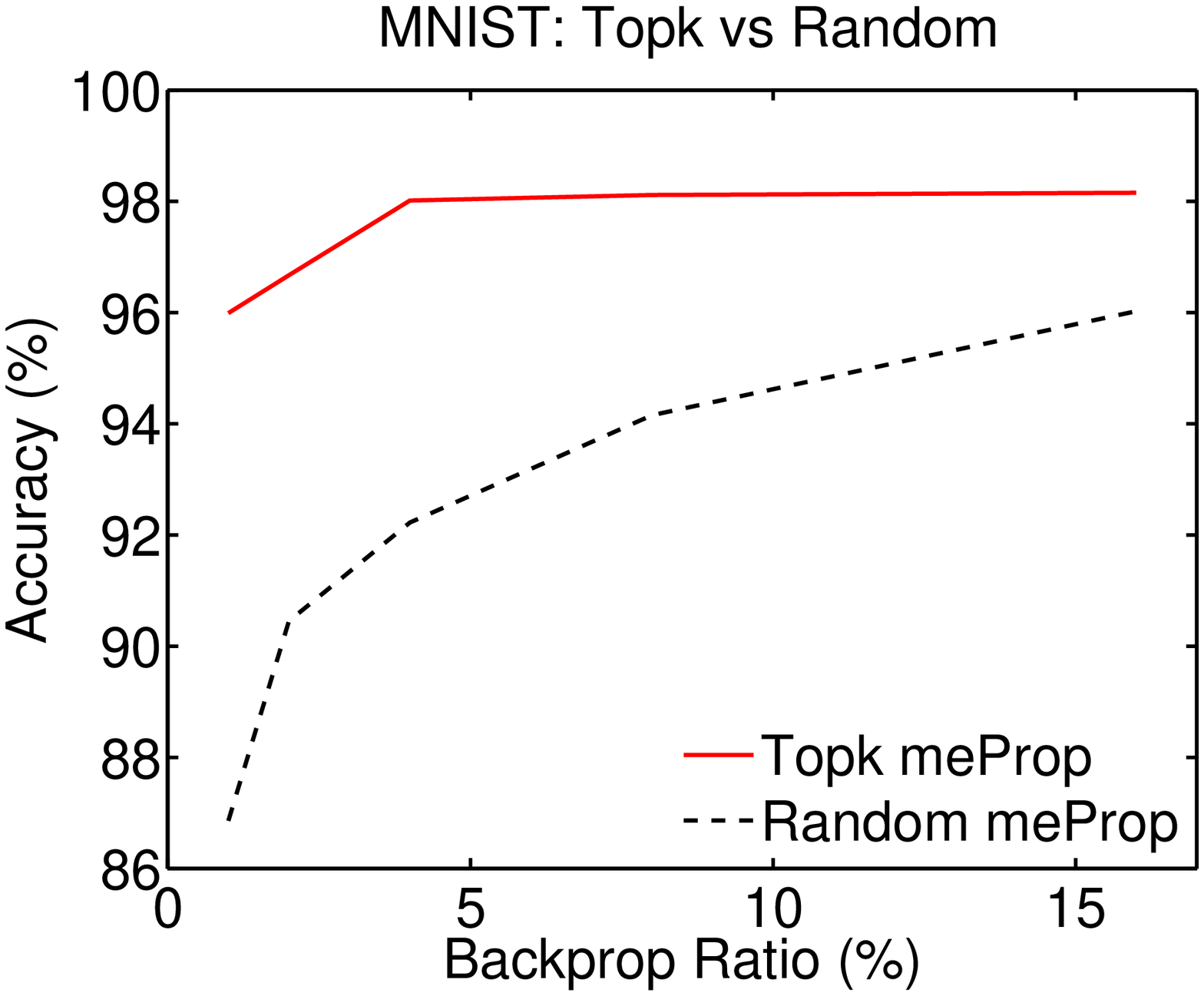,width=0.3\linewidth,clip=}
\epsfig{file=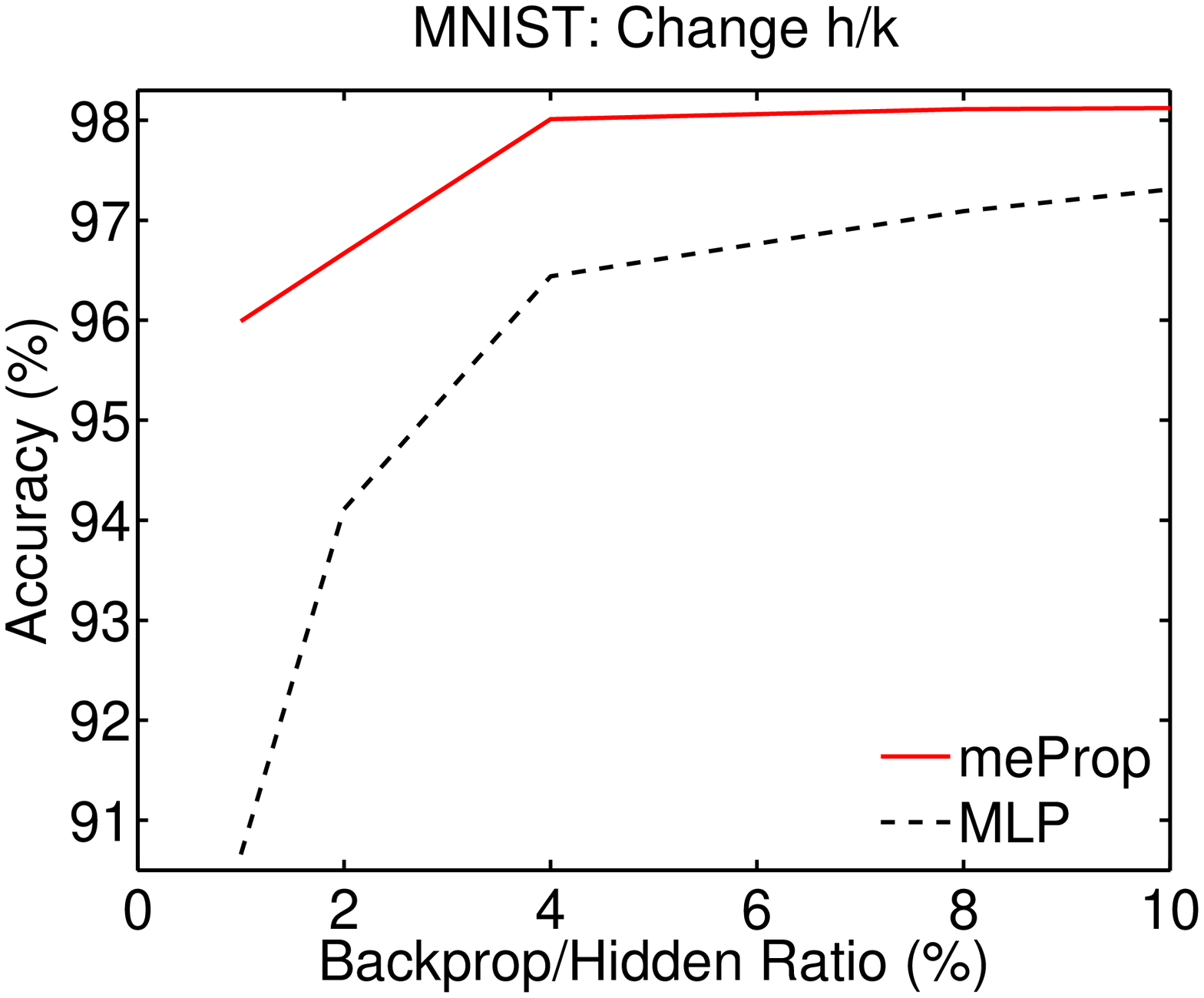,width=0.3\linewidth,clip=} \\

\end{tabular}
\caption{Accuracy vs. meProp's backprop ratio (left). Results of top-$k$ meProp vs. random meProp (middle). Results of top-$k$ meProp vs. baseline with the hidden dimension $h$ (right).
}\label{fig4}
\vspace{-0.05in}
\end{figure*}

To demonstrate that the proposed method is general-purpose, we perform experiments on different models (LSTM/MLP), various training methods (Adam/AdaGrad), and diverse tasks.

\textbf{Part-of-Speech Tagging (POS-Tag):}
We use the standard benchmark dataset in prior work \cite{Collins2002}, which is derived from the Penn Treebank corpus, and use sections 0-18 of the Wall Street Journal (WSJ) for training (38,219 examples), and sections 22-24 for testing (5,462 examples).
The evaluation metric is per-word accuracy. A popular model for this task is the LSTM model \cite{Hochreiter1997},\footnote{In this work, we use the bi-directional LSTM (Bi-LSTM) as the implementation of LSTM.} which is used as our baseline.

\textbf{Transition-based Dependency Parsing (Parsing):}
Following prior work, we use English Penn TreeBank (PTB) \cite{Marcusetal1993} for evaluation. We follow the standard split of the corpus and use sections 2-21 as the training set (\num{39832} sentences, \num{1900056} transition examples),\footnote{A transition example consists of a parsing context and its optimal transition action.} section 22 as the development set (\num{1700} sentences, \num{80234} transition examples) and section 23 as the final test set (\num{2416} sentences, \num{113368} transition examples).
The evaluation metric is \emph{unlabeled attachment score} (UAS). We implement a parser using MLP following Chen and Manning \yrcite{Chen2014}, which is used as our baseline.

\textbf{MNIST Image Recognition (MNIST):}
We use the MNIST handwritten digit dataset \cite{lecun1998gradient} for evaluation. MNIST consists of \num{60000} 28$\times$28 pixel training images and additional \num{10000} test examples. Each image contains a single numerical digit (0-9). We select the first \num{5000} images of the training images as the development set and the rest as the training set.
The evaluation metric is per-image accuracy. We use the MLP model as the baseline.

\subsection{Experimental Settings}

We set the dimension of the hidden layers to 500 for all the tasks.
For POS-Tag, the
input dimension is $1\text{ (word)}\times50\text{ (dim per word)}+7\text{ (features)}\times20\text{ (dim per feature)}=190$, and
the output dimension is $45$.
For Parsing, the
input dimension is $48\text{ (features)}\times50\text{ (dim per feature)}=2400$,
and the output dimension is $25$.
For MNIST, the
input dimension is $28\text{ (pixels per row)}\times28\text{ (pixels per column)}\times1\text{ (dim per pixel)}=784$,
and the output dimension is $10$.
As discussed in Section 2, the optimal $k$ of top-$k$ for the output layer could be different from the hidden layers, because their dimensions could be very different. For Parsing and MNIST, we find using the same $k$ for the output and the hidden layers works well, and we simply do so. For another task, POS-Tag, we find the the output layer should use a different $k$ from the hidden layers. For simplicity, we do not apply meProp to the output layer for POS-Tag, because in this task we find the computational cost of the output layer is almost negligible compared with other layers.

The hyper-parameters are tuned based on the development data.
For the Adam optimization method, we find the default hyper-parameters work well on development sets, which are as follows: the learning rate $\alpha=0.001$, and $\beta_1=0.9, \beta_2=0.999$, $\epsilon=1\times10^{-8}$.
For the AdaGrad learner, the learning rate is set to $\alpha=0.01, 0.01, 0.1$ for POS-Tag, Parsing, and MNIST, respectively, and $\epsilon=1\times10^{-6}$.
The experiments on CPU are conducted on a computer with the \emph{INTEL(R) Xeon(R) 3.0GHz CPU}. The experiments on GPU are conducted on \emph{NVIDIA GeForce GTX 1080}.

\subsection{Experimental Results}

In this experiment, the LSTM is based on one hidden layer and the MLP is based on two hidden layers (experiments on more hidden layers will be presented later).
We conduct experiments on different optimization methods, including AdaGrad and Adam. Since meProp is applied to the linear transformations (which entail the major computational cost), we report the linear transformation related backprop time as \textit{Backprop Time}. It does not include non-linear activations, which usually have only less than 2\% computational cost. The total time of back propagation, including non-linear activations, is reported as \textit{Overall Backprop Time}.
Based on the development set and prior work, we set the mini-batch size to 1 (sentence), \num{10000} (transition examples), and 10 (images) for POS-Tag, Parsing, and MNIST, respectively. Using \num{10000} transition examples for Parsing follows Chen and Manning \yrcite{Chen2014}.

Table \ref{tab1} shows the results based on different models and different optimization methods.
In the table, \emph{meProp} means applying meProp to the corresponding baseline model, $h=500$ means that the hidden layer dimension is 500, and $k=20$ means that meProp uses top-20 elements (among 500 in total) for back propagation. Note that, for fair comparisons, all experiments are first conducted on the development data and the test data is not observable. Then, the optimal number of iterations is decided based on the optimal score on development data, and the model of this iteration is used upon the test data to obtain the test scores.

As we can see, applying meProp can substantially speed up the back propagation. It provides a linear reduction in the computational cost. Surprisingly, results demonstrate that we can update only 1--4\% of the weights at each back propagation pass. This does not result in a larger number of training iterations. More surprisingly, the accuracy of the resulting models is actually improved rather than decreased.
The main reason could be that the minimal effort update does not modify weakly relevant parameters, which makes overfitting less likely, similar to the \emph{dropout} effect.

Table \ref{tab2} shows the overall forward propagation time, the overall back propagation time, and the training time by summing up forward and backward propagation time. As we can see, back propagation has the major computational cost in training LSTM/MLP.

The results are consistent among AdaGrad and Adam. The results demonstrate that meProp is independent of specific optimization methods. For simplicity, in the following experiments the optimizer is based on Adam.


\begin{table}[t]
  \centering
    \footnotesize
  \caption{Results based on the same $k$ and $h$.} \label{tab3}
  \begin{tabular}{|l|r|l|}
    \hline
    \textbf{POS-Tag (Adam)} & Iter & Test Acc (\%)\\
    \hline
     LSTM (h=5) & 7  & 96.40  \\

     meProp (k=5) & 5  & \textbf{97.12 (+0.72)} \\
    \hline

    \textbf{Parsing (Adam)} & Iter & Test UAS (\%)\\
    \hline
     MLP (h=20) & 18  & 88.37  \\

     meProp (k=20) & 6 & \textbf{90.01 (+1.64)} \\
    \hline

    \textbf{MNIST (Adam)} & Iter & Test Acc (\%)\\
    \hline
     MLP (h=20) & 15  & 95.77   \\
 
     meProp (k=20) & 17 & \textbf{98.01 (+2.24)} \\
    \hline
  \end{tabular}
  \vspace{-0.05in}
\end{table}

\subsection{Varying Backprop Ratio}

In Figure \ref{fig4} (left), we vary the $k$ of top-$k$ meProp to compare the test accuracy on different ratios of meProp backprop. For example, when k=5, it means that the backprop ratio is 5/500=1\%. The optimizer is Adam. As we can see, meProp achieves consistently better accuracy than the baseline. The best test accuracy of meProp, 98.15\% (+0.33), is actually better than the one reported in Table \ref{tab1}.

\subsection{Top-k vs. Random}

It will be interesting to check the role of top-$k$ elements.
Figure \ref{fig4} (middle) shows the results of top-$k$ meProp vs. random meProp. The \emph{random meProp} means that random elements (instead of top-$k$ ones) are selected for back propagation.
As we can see, the top-$k$ version works better than the random version. It suggests that top-$k$ elements contain the most important information of the gradients. 

\subsection{Varying Hidden Dimension}

We still have a question: does the top-$k$ meProp work well simply because the original model does not require that big dimension of the hidden layers? For example, the meProp (topk=5) works simply because the LSTM works well with the hidden dimension of 5, and there is no need to use the hidden dimension of 500. To examine this, we perform experiments on using the same hidden dimension as $k$, and the results are shown in Table \ref{tab3}. As we can see, however, the results of the small hidden dimensions are much worse than those of meProp.

In addition, Figure \ref{fig4} (right) shows more detailed curves by varying the value of $k$. In the figure, different $k$ gives different backprop ratio for meProp and different hidden dimension ratio for LSTM/MLP. As we can see, the answer to that question is negative: meProp does not rely on redundant hidden layer elements.


\begin{table}[t]
  \centering
    \footnotesize
  \caption{Adding the \emph{dropout} technique. } \label{tab4}
  \begin{tabular}{|l|r|l|}
    \hline

    \textbf{POS-Tag (Adam)} & Dropout  & Test Acc (\%)\\
    \hline
     LSTM (h=500) & 0.5  & 97.20 \\

     meProp (k=20) & 0.5  &  \textbf{97.31 (+0.11)} \\
    \hline

    \textbf{Parsing (Adam)} & Dropout  & Test UAS (\%)\\
    \hline
     MLP (h=500) & 0.5 & 91.53 \\
     meProp (k=40) & 0.5 &  \textbf{91.99 (+0.46)} \\
    \hline

    \textbf{MNIST (Adam)} & Dropout  & Test Acc (\%)\\
    \hline
     MLP (h=500) & 0.2 & 98.09 \\
     meProp (k=25) & 0.2 &  \textbf{98.32 (+0.23)} \\
    \hline
  \end{tabular}
  \vspace{-0.05in}
\end{table}


\begin{table}[t]
  \centering
    \footnotesize
  \caption{Varying the number of hidden layers on the MNIST task. The optimizer is Adam. Layers: the number of hidden layers.} \label{tab5}
  \begin{tabular}{|c|l|l|}
    \hline

    Layers & Method & Test Acc (\%) \\
    \hline

    \multirow{2}{*}{2} & MLP (h=500) & 98.10   \\

                       & meProp (k=25)  & \textbf{98.20 (+0.10)} \\
    \hline

    \multirow{2}{*}{3} & MLP (h=500)  & 98.21   \\

                       & meProp (k=25)  & \textbf{98.37 (+0.16)} \\
    \hline

    \multirow{2}{*}{4} & MLP (h=500)  & 98.10   \\

                       & meProp (k=25)  & \textbf{98.15 (+0.05)} \\
    \hline

    \multirow{2}{*}{5} & MLP (h=500)  & 98.05   \\

                       & meProp (k=25)  & \textbf{98.21 (+0.16)} \\
    \hline
  \end{tabular}
  \vspace{-0.05in}
\end{table}

\subsection{Adding Dropout}

Since we have observed that meProp can reduce overfitting of deep learning, a natural question is that if meProp is reducing the same type of overfitting risk as dropout. Thus, we use development data to find a proper value of the dropout rate on those tasks, and then further add meProp to check if further improvement is possible.

Table~\ref{tab4} shows the results. As we can see, meProp can achieve further improvement over dropout. In particular, meProp has an improvement of 0.46 UAS on Parsing. The results suggest that the type of overfitting that meProp reduces is probably different from that of dropout. Thus, a model should be able to take advantage of both meProp and dropout to reduce overfitting.

\subsection{Adding More Hidden Layers }

Another question is whether or not meProp relies on shallow models with only a few hidden layers. To answer this question,
we also perform experiments on more hidden layers, from 2 hidden layers to 5 hidden layers. We find setting the dropout rate to 0.1 works well for most cases of different numbers of layers. For simplicity of comparison, we set the same dropout rate to 0.1 in this experiment. Table~\ref{tab5} shows that adding the number of hidden layers does not hurt the performance of meProp.

\begin{table}[t]
  \centering
    \footnotesize
  \caption{Results of \emph{simple unified top-$k$} meProp based on a whole mini-batch (i.e., unified sparse patterns). The optimizer is Adam. Mini-batch Size is 50.} \label{tab6}

  \begin{tabular}{|c|l|l|}
    \hline
    Layers & Method  & Test Acc (\%) \\
    \hline

    \multirow{2}{*}{2} & MLP (h=500)  & 97.97   \\

                       & meProp (k=30) & \textbf{98.08 (+0.11)} \\
    \hline

    \multirow{2}{*}{5} & MLP (h=500)  & 98.09   \\

                       & meProp (k=50)  & \textbf{98.36 (+0.27)} \\
    \hline
  \end{tabular}
  \vspace{-0.05in}
\end{table}


\begin{table}[t]
  \centering
    \footnotesize
  \caption{Acceleration results on the matrix multiplication synthetic data using GPU. The batch size is 1024.} \label{tab7}
  \begin{tabular}{|l|r|}
    \hline
    Method & Backprop time (ms) \\
    \hline
    Baseline (h=8192) & \num{308.00}\hspace{28.5pt} \\
    \hline

    meProp (k=8) & \num{8.37} (36.8x) \\
    \hline

    meProp (k=16) & \num{9.16} (33.6x) \\
    \hline

    meProp (k=32) & \num{11.20} (27.5x) \\
    \hline

    meProp (k=64) & \num{14.38} (21.4x) \\
    \hline

    meProp (k=128) & \num{21.28} (14.5x) \\
    \hline

    meProp (k=256) & \num{38.57} (8.0x)\hspace{5pt} \\
    \hline

    meProp (k=512) & \num{69.95} (4.4x)\hspace{5pt} \\
    \hline
  \end{tabular}
  \vspace{-0.05in}
\end{table}


\begin{table}[t]
  \centering
    \footnotesize
  \caption{Acceleration results on MNIST using GPU.} \label{tab8}
  \begin{tabular}{|l|r|}
    \hline
    Method & Overall backprop time (ms) \\
    \hline

    MLP (h=8192) & \num{17696.2}\hspace{28.5pt} \\
    \hline

    meProp (k=8) & \num{1501.5} (11.8x) \\
    \hline

    meProp (k=16) & \num{1542.8} (11.5x) \\
    \hline

    meProp (k=32) & \num{1656.9} (10.7x) \\
    \hline

    meProp (k=64) & \num{1828.3} (9.7x)\hspace{5pt} \\
    \hline

    meProp (k=128) & \num{2200.0} (8.0x)\hspace{5pt} \\
    \hline

    meProp (k=256) & \num{3149.6} (5.6x)\hspace{5pt} \\
    \hline

    meProp (k=512) & \num{4874.1} (3.6x)\hspace{5pt} \\
    \hline
  \end{tabular}
  \vspace{-0.05in}
\end{table}

\subsection{Speedup on GPU}

For implementing meProp on GPU, the simplest solution is to treat the entire mini-batch as a ``big training example'', where the top-$k$ operation is based on the averaged values of all examples in the mini-batch. In this way, the big sparse matrix of the mini-batch will have consistent sparse patterns among examples, and this consistent sparse matrix can be transformed into a small dense matrix by removing the zero values. We call this implementation as \emph{simple unified top-$k$}. This experiment is based on \emph{PyTorch}.

Despite its simplicity, Table~\ref{tab6} shows the good performance of this implementation, which is based on the mini-batch size of 50. We also find the speedup on GPU is less significant when the hidden dimension is low. The reason is that our GPU's computational power is not fully consumed by the baseline (with small hidden layers), so that the normal back propagation is already fast enough, making it hard for meProp to achieve substantial speedup. For example, supposing a GPU can finish 1000 operations in one cycle, there could be no speed difference between a method with 100 and a method with 10 operations. Indeed, we find MLP (h=64) and MLP (h=512) have almost the same GPU speed even on \emph{forward} propagation (i.e., without meProp), while theoretically there should be an 8x difference. With GPU, the \emph{forward} propagation time of MLP (h=64) and MLP (h=512) is 572ms and 644ms, respectively. This provides evidence for our hypothesis that our GPU is not fully consumed with the small hidden dimensions.

Thus, the speedup test on GPU is more meaningful for the heavy models, such that the baseline can at least fully consume the GPU's computational power.
To check this, we test the GPU speedup on synthetic data of matrix multiplication with a larger hidden dimension. Indeed, Table~\ref{tab7} shows that meProp achieves much higher speed than the traditional backprop with the large hidden dimension. Furthermore, we test the GPU speedup on MLP with the large hidden dimension \cite{DrydenEA2016}. Table~\ref{tab8} shows that meProp also has substantial GPU speedup on MNIST with the large hidden dimension. In this experiment, the speedup is based on \emph{Overall Backprop Time} (see the prior definition). Those results demonstrate that meProp can achieve good speedup on GPU when it is applied to heavy models.

Finally, there are potentially other implementation choices of meProp on GPU. For example, another natural solution is to use a big sparse matrix to represent the sparsified gradient of the output of a mini-batch. Then, the sparse matrix multiplication library can be used to accelerate the computation. This could be an interesting direction of future work.

\subsection{Related Systems on the Tasks}

The POS tagging task is a well-known benchmark task, with the accuracy reports from 97.2\% to 97.4\% \cite{ToutanovaKMS03,Sun_NIPS2014,acl/ShenSJ07,Tsuruoka2011,CollobertEA2011,huang2015bidirectional}. Our method achieves 97.31\% (Table~\ref{tab4}).

For the transition-based dependency parsing task, existing approaches typically can achieve the UAS score from 91.4 to 91.5~\cite{Zhang2008,Nirve2007,Huang2010}. As one of the most popular transition-based parsers, MaltParser~\cite{Nirve2007} has 91.5 UAS. Chen and Manning~\yrcite{Chen2014} achieves 92.0 UAS using neural networks. Our method achieves 91.99 UAS (Table~\ref{tab4}).

For MNIST, the MLP based approaches can achieve 98--99\% accuracy, often around 98.3\% \cite{lecun1998gradient, simardEA2003, ciresanEA2010}. Our method achieves 98.37\% (Table~\ref{tab5}). With the help from convolutional layers and other techniques, the accuracy can be improved to over 99\% \cite{JarrettEA2009, CiresanEA2012}. Our method can also be improved with those additional techniques, which, however, are not the focus of this paper.

\section{Conclusions}

The back propagation in deep learning tries to modify all parameters in each stochastic update, which is inefficient and may even lead to overfitting due to unnecessary modification of many weakly relevant parameters.
We propose a \emph{minimal effort} back propagation method (\emph{meProp}), in which we compute only a very small but critical portion of the gradient, and modify only the corresponding small portion of the parameters in each update. This leads to very sparsified gradients to modify only highly relevant parameters for the given training sample. The proposed meProp is independent of the optimization method. Experiments show that meProp can reduce the computational cost of back propagation by one to two orders of magnitude via updating only 1--4\% parameters, and yet improve the model accuracy in most cases.

\section*{Acknowledgements}

The authors would like to thank the anonymous reviewers for insightful comments and suggestions on this paper.
This work was supported in part by National Natural Science Foundation of China (No. 61673028), National High Technology Research and Development Program of China (863 Program, No. 2015AA015404), and an Okawa Research Grant (2016).

\end{document}